\documentclass[journal]{IEEEtran}
\usepackage[T1]{fontenc}

\usepackage{graphicx}
\usepackage{epstopdf}

\ifCLASSINFOpdf
\else
\fi
\usepackage{amsmath}
\usepackage{array}
\usepackage{amsmath,amssymb}

\begin{document}
%
\title{Automatic Image Labelling at Pixel Level}
%
%
%

\author{Xiang~Zhang,
        Wei~Zhang,
        Jinye~Peng,
        Jianping~Fan
\thanks{This work has been submitted to the IEEE for possible publication. Copyright may be transferred without notice, after which this version may no longer be accessible.}
\thanks{Xiang Zhang is with the School of Information and Technology, Northwest University, Shaanxi 710127, China (e-mail: ZhangXiang2015@stumail.nwu.edu.cn).}
\thanks{Wei Zhang is with the School of Computer Science, Fudan University, Shanghai 200433, China (e-mail: weizh@fudan.edu.cn).}
\thanks{Jinye Peng is with the School of Information and Technology, Northwest University, Shaanxi 710127, China (e-mail: pjy@nwu.edu.cn).}
\thanks{Jianping Fan is with the Department of Computer Science, The University of North Carolina at Charlotte, Charlotte, NC 28223 USA (e-mail: jfan@uncc.edu).}}

%
%

\markboth{SUBMITTED TO IEEE TRANSACTION ON IMAGE PROCESSING, 2020}%
{Zhang \MakeLowercase{\textit{\emph{et al.}}}: Automatic Image Labelling at Pixel Level}
%



\maketitle


\begin{abstract}
The performance of deep networks for semantic image segmentation largely depends on the availability of large-scale training images which are labelled at the pixel level. Typically, such pixel-level image labellings are obtained manually by a labour-intensive process. To alleviate the burden of manual image labelling, we propose an interesting learning approach to generate pixel-level image labellings automatically. A Guided Filter Network (GFN) is first developed to learn the segmentation knowledge from a source domain, and such GFN then transfers such segmentation knowledge to generate coarse object masks in the target domain. Such coarse object masks are treated as pseudo labels and they are further integrated to optimize/refine the GFN iteratively in the target domain. Our experiments on six image sets have demonstrated that our proposed approach can generate fine-grained object masks (i.e., pixel-level object labellings), whose quality is very comparable to the manually-labelled ones. Our proposed approach can also achieve better performance on semantic image segmentation than most existing weakly-supervised approaches.
\end{abstract}



%
\IEEEpeerreviewmaketitle

\begin{figure*}[t]
\begin{center}
\includegraphics[width=0.65\linewidth]{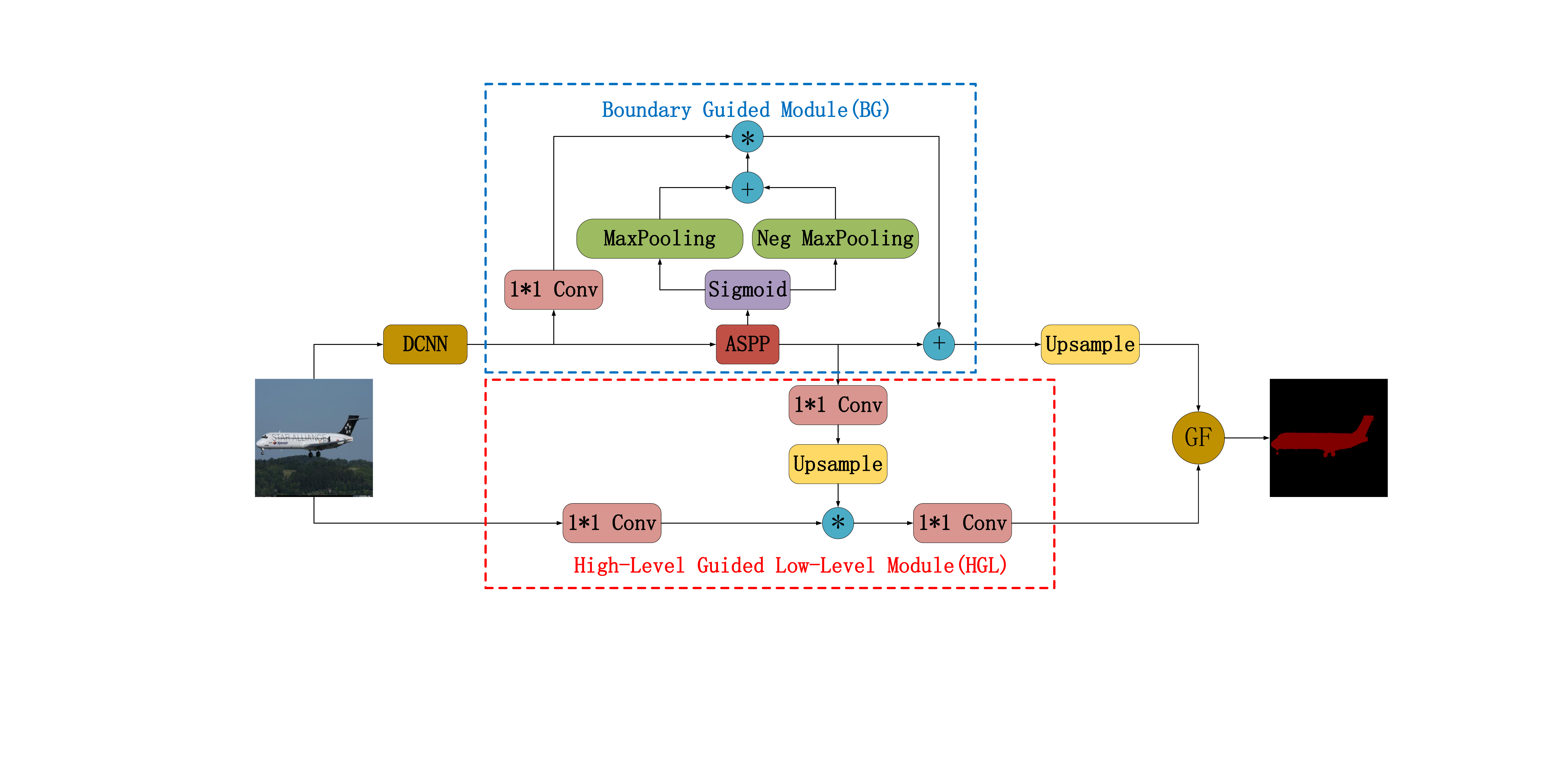}
\end{center}
\vspace*{-0.38cm}
  \caption{Overview of our proposed GFN Network. DCNN (ResNet-101) is employed to extract dense features, BG is performed to recover localization details and HGL is used to filter the low-level noise features. Lastly,  Guided Filter (GF) is designed to recover object structure and extract the precise results. $+$ and $*$ represent the element-wise addition and elements-wise multiplication, respectively.}
\label{fig:network}
\vspace*{-0.28cm}
\end{figure*}

\section{Introduction}

\IEEEPARstart{S}{emantic} image segmentation, which assigns one particular semantic label to each pixel in an image, is a critical task in computer vision. Recently, Deep Convolutional Neural Networks (DCNNs) have demonstrated their outstanding performance on the task of semantic image segmentation~\cite{Long_2015_CVPR,chen2017deeplab,chen2018encoder}. Obviously, their success is largely owed to the availability of large-scale training images whose semantic labels are given precisely at the pixel level. However, manually labelling large-scale training images at the pixel level is dreadfully labour-intensive and it could also be very difficult for humans to provide consistent labelling quality.

To alleviate the burden of manual image labelling, some weakly-supervised approaches have been developed to support deep semantic image segmentation. Instead of requiring precise pixel-level image labellings, such weakly-supervised methods employ the semantic labels that are coarsely given at the level of image bounding boxes~\cite{Dai_2015_ICCV} or image scribbles~\cite{Lin_2016_CVPR}. To further reduce human involvements in image labelling, numerous methods~\cite{kolesnikov2016seed,wei2017object,zhang2019decoupled} train the deep networks from large-scale training images with coarse image-level labellings. Unfortunately, the performance of all these weakly-supervised methods is far from satisfactory because pixel-level image labellings play a critical role on learning discriminative networks for semantic  segmentation. Thus it is very attractive to develop new algorithms that are able to automatically generate fine-grained object masks with detailed pixel-level structures/boundaries (i.e., pixel-level object labellings), so that we can learn more discriminative networks for deep semantic image segmentation.

Recently, many self-supervised learning methods \cite{noroozi2016unsupervised,zhang2016colorful,pathak2016context} were proposed to learn visual features from large-scale unlabelled images by using their pseudo labels rather than using human annotations. This new technique has been widely applied to the task of object recognition \cite{zhai2019s4l,kolesnikov2019revisiting,misra2020self}, object detection \cite{he2020momentum} and image segmentation \cite{larsson2016learning,novoselboosting}. The main idea is to design a pre-defined pretext task for DCNNs to solve, and the pseudo labels for the pretext task can be automatically generated by using attributes of data. Then the
DCNNs is trained to learn object functions of the pretext task. After the self-supervised training finished, the learned visual features can be further transferred to downstream tasks as initialization parameters to improve performance \cite{jing2020self}.

Based on these observations, an interesting approach, called Guided Filter Network (GFN), is developed in this paper to automatically generate pixel-level image labellings. The proposed approach can be divided into two stages: the proxy stage, and the iterative learning stage. The proxy stage does not need any pixel-level image labellings in the target domain but requires one to design a pretext task to learn a segmentation knowledge in the source domain. In the iterative learning stage, the learned parameters are transferred to GFN for generating initial object masks for the images in the target domain. Such coarse object masks are further leveraged to optimiza/refine the GFN iteratively in the target domain. For a given object class of interest, we designed two strategies to obtain its source domain: (a) if itself or its semantically-similar category (i.e., its sibling classes under the same parent node, or its parents or its ancestors on a concept ontology) can be identified from public datasets (which provide pixel-level object labellings), e.g. PASCAL VOC~\cite{everingham2015pascal}, Microsoft COCO~\cite{lin2014microsoft} and BSD~\cite{russell2008labelme}, we can treat such semantically-similar category and its pixel-level labelled images in public datasets as its source domain; (b) if itself or its semantically-similar category cannot be identified from public datasets, we can first search some relevant images with simple backgrounds from the Internet and an Otsu detector~\cite{otsu1979threshold} is utilized to extract the given object class of interest and generate pixel-level object labellings automatically.

Our proposed GFN contains three modules: Boundary Guided (BG) module, Guided Filter (GF) module and High-level Guided Low-level (HGL) module. GFN can recover the structures and  the localization details of objects effectively. For a given object class of interest, GFN first learns a segmentation knowledge from the source domain. GFN then transfers such segmentation knowledge to generate coarse object masks from the images in the target domain. Those coarse object masks are treated as pseudo labels and they are further leveraged to optimize/refine the GFN iteratively, so that it is able to generate the fine-grained object masks with detailed pixel-level structures/boundaries (i.e., pixel-level object labellings) gradually.

\section{Related Work}
Deep Convolutional Neural Networks (DCNNs) have achieved a great success on semantic image segmentation. One of the most popular CNN-based approaches is the Fully Convolution Network (FCN)~\cite{Long_2015_CVPR}. Chen \emph{et al.}~\cite{chen2014semantic} improves the FCN-style architecture by applying fully connected CRF as a post-processing step. Several deep models are proposed to exploit the contextual information for semantic image segmentation \cite{chen2017deeplab,yu2015multi}. Other methods~\cite{chen2018encoder,ronneberger2015u} have built a deep encoder-decoder structure to preserve object boundaries precisely. Although these existing approaches have achieved a substantial improvement on deep semantic image segmentation, a critical bottleneck is to collect large amounts of training images with pixel-level labellings and this bottleneck may seriously limit their practicality~\cite{kwak2017weakly}.

To tackle the deficiency of training images with pixel-level labellings, some weakly-supervised approaches are proposed, where deep models are learned from large-scale training images whose semantic labels are given coarsely at the level of bounding boxes or scribbles rather than at the pixel level. For example, Dai \emph{et al.}~\cite{Dai_2015_ICCV} proposed to learn from the training images with coarse object masks at the level of bounding boxes. The basic idea is to perform an iteration process between automatically generating region proposals and training convolutional networks. Lin \emph{et al.}~\cite{Lin_2016_CVPR} used the scribbles to label the training images for network training.

To further reduce human involvements in manual image labelling, some deep models directly learn from large-scale training images whose semantic labels are coarsely given at the image level. By adding an extra layer for multiple-instance learning (MIL), Pinheiro \emph{et al.}~\cite{Pinheiro_2015_CVPR} learned deep models which are able to assign more weights to the important pixels. Kolesnikov \emph{et al.}~\cite{kolesnikov2016seed} proposed to learn the segmentation network from the classification networks. Some researches~\cite{wei2017object,zhang2019decoupled} used the training images with image-level labellings to learn the semantic segmentation networks. Unfortunately, the performance of all these weakly-supervised methods is far from satisfactory. One possible reason is that the learned segmentation networks may not be able to recover the original spatial structures of objects precisely because the intrinsic properties of pixel-level object masks are completely ignored.

Different from these methods, for a given object class of interest, our proposed approach takes the following steps to achieve deep semantic image segmentation: (a) the GFN first generates coarse object masks automatically by transferring the segmentation knowledge learned from the source domain to the target domain; (b) such coarse object masks are further leveraged to optimize/refine the GFN iteratively in the target domain, so that we can obtain fine-grained object masks with detailed pixel-level structures/boundaries gradually.

\begin{figure*}[t]
\begin{center}
\includegraphics[width=0.69\linewidth]{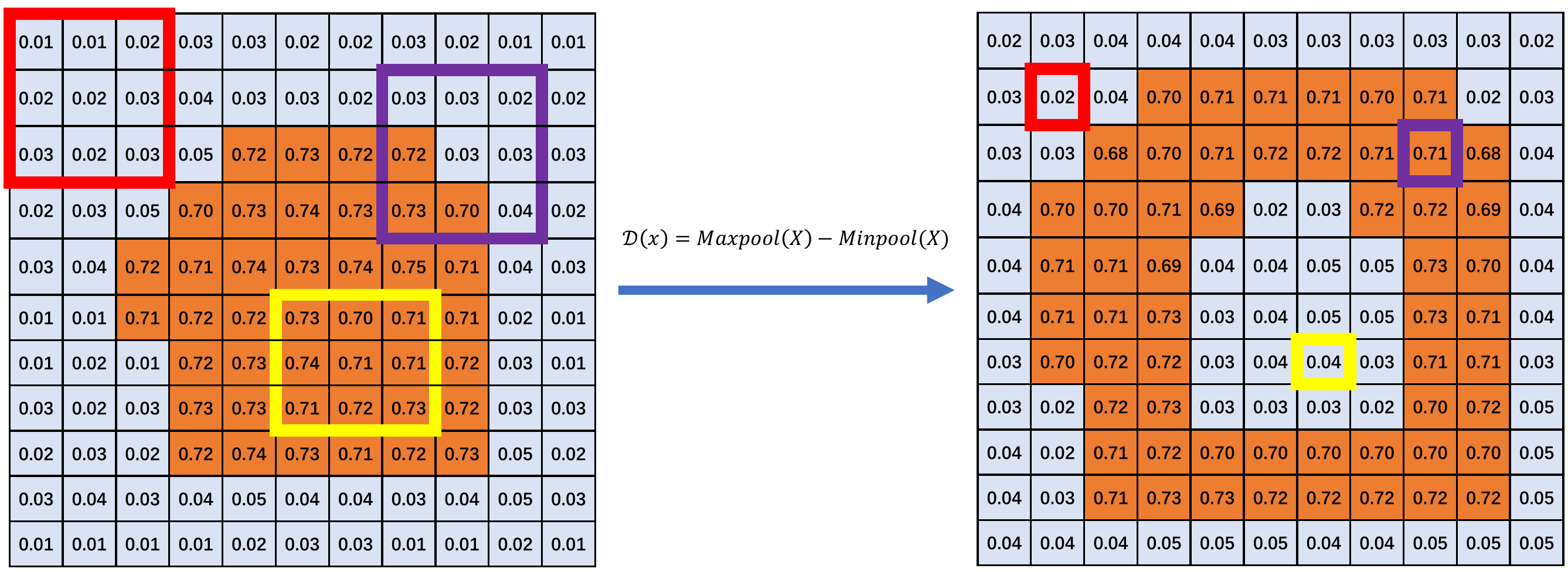}
\end{center}
\vspace*{-0.38cm}
  \caption{Boundary extractor. A example of boundary extraction operate with a kernel size of $3\times3$, a padding of 1, and a stride of 1.}
\label{fig:bgm_image}
\vspace*{-0.28cm}
\end{figure*}

\section{Architecture of our Proposed GFN} 
In this section, we first introduce three components of our proposed Guided Filter Network (GFN): (a) Boundary Guided (BG) module; (b) Guided Filter (GF) module; and (c) High-level Guided Low-level (HGL) module. Then we elaborate our complete GFN architecture.

\subsection{Boundary Guided Module}
In the task of semantic segmentation, numerous method~\cite{chen2018encoder,ronneberger2015u} usually combine low-level and high-level features to boost performance. Compared to simply fusing low-level features with high-level ones, Zhang \emph{et al.}~\cite{zhang2018exfuse} pointed out that later fusion is more effective by introducing high-resolution details into high-level features and embedding semantic information into low-level features.

Inspired by the works in~\cite{zhang2018exfuse,ghiasi2016laplacian}, we establish a Boundary Guided (BG) module that can recover the localization details of objects by introducing low-level features into high-level ones and embedding semantic information into low-level features. The structure of BG module is shown in Figure~\ref{fig:network}, where a boundary extractor is first applied to obtain the semantic features of object boundaries. Based on the observation that the pixel values in the non-boundary areas change slowly while the pixel values around the boundary change significantly, the boundary extractor can be designed as:
\begin{equation}
   \mathcal{D}(X) = Maxpool(X) - Minpool(X)
   \label{eq:boundex}
 \end{equation}
where $X$ represents the feature maps,  $Maxpool(.)$ and $Minpool(.)$ are max-pooling and min-pooling operations respectively. In practice Eq. (\ref{eq:boundex}) can be implemented by  $\mathcal{D}(X) = Maxpool(X) + Maxpool(-X)$. Thus, the boundary extractor makes that the values of pixels not around the non-boundary are close to zero while others are not (see Figure~\ref{fig:bgm_image}). After that, an element-wise multiplication is performed between the low-level features and the semantic features of object boundaries.

\begin{figure*}[t]
\begin{center}
\includegraphics[width=0.9\linewidth]{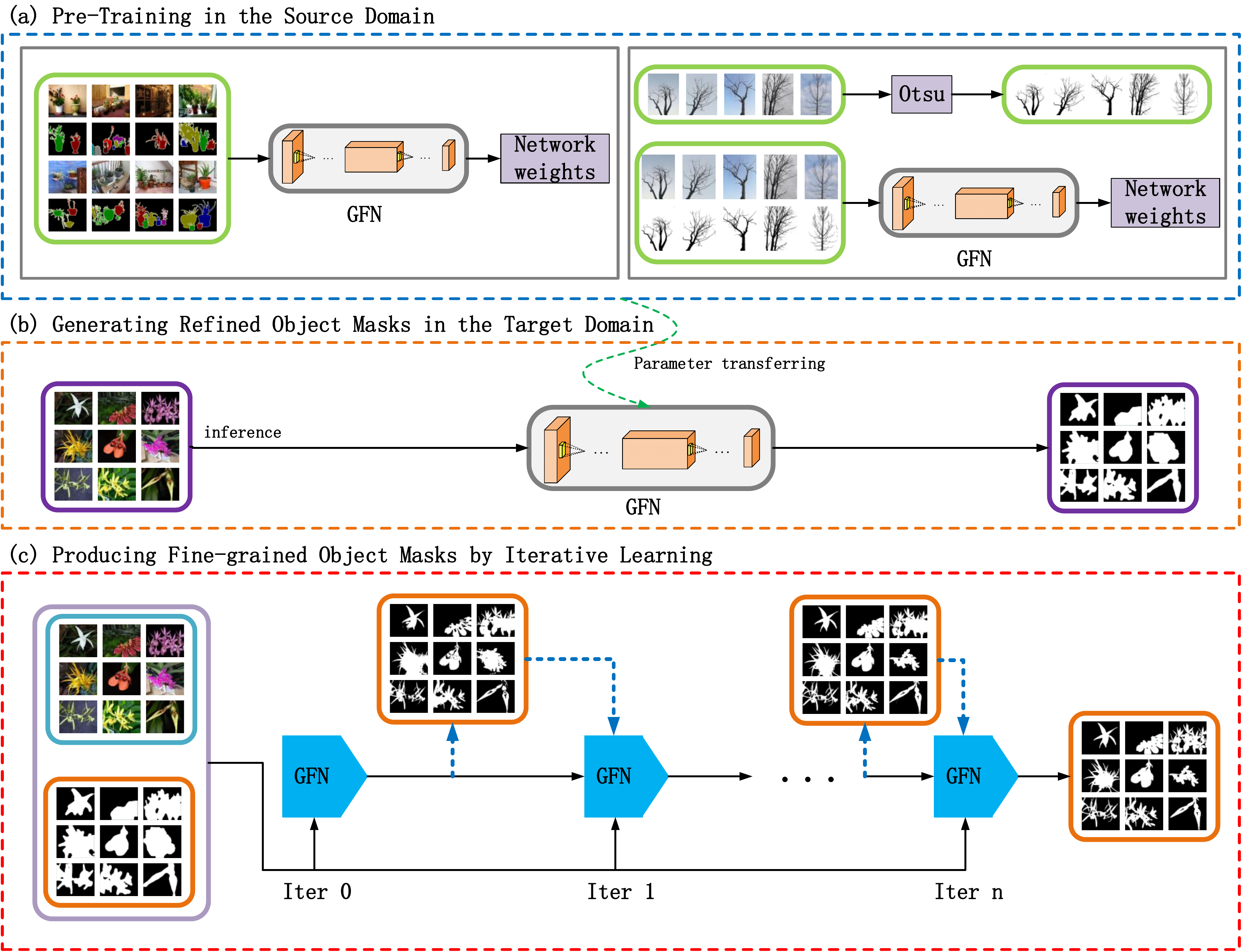}
\end{center}
\vspace*{-0.38cm}
  \caption{The overview of our proposed approach.}
\label{fig:framework}
\vspace*{-0.28cm}
\end{figure*}

\subsection{Guided Filter Module}
DCNNs have demonstrated their outstanding capability on the task of semantic image segmentation, but they have difficulty in obtaining good results with detailed pixel-level structures/boundaries. One core issue is that obtaining object-centric decisions from a classifier requires the invariance to spatial transformations, which inherently limits the spatial accuracy of the DCNNs~\cite{chen2017deeplab}. Several methods have recently been proposed to alleviate this issue and they can be grouped into two main strategies. The first strategy is to build deep encoder-decoder networks which can preserve accurate object boundaries~\cite{chen2018encoder,ronneberger2015u}. The second strategy is to use CRF to refine the segmentation results iteratively~\cite{chen2017deeplab,Papandreou_2015_ICCV}. Although these approaches have achieved substantial improvements on semantic image segmentation, using multiple down-sampling layers may lead to more information loss.

The guided filter~\cite{he2013guided} is an edge-preserving operator and can transfer the structure of the guidance image (e.g., original RGB image or low-level features) to the filtering output. In this paper,   we use the guided filter to extract the edge contour coefficient for each pixel from the guidance image, then such coefficients are used to weight the coarse object masks for generating the refined ones. More specifically, suppose that two inputs, the coarse object mask (e.g., the output of BG module) $p$ and the guidance image $I$, are fed to the guided filtering. The outputs of the guided filter (i.e., the refined masks) are obtained as follows:

 \begin{equation}
   g_{i} = \sum_{j}W_{ij}(I)p_{j}, \hspace*{0.2cm} \forall i \in w_{k}
 \label{eq:output}
 \end{equation}
where $i$ is the index of a pixel in $I$, $j$ is the index of a point in $p$, and  $k$ is the index of a local square window $w$ with radius $r$. $W_{ij}$  is the edge contour coefficient of the pixel $(i, j)$ in the image $I$ and it is calculated as:
\begin{equation}
    W_{ij}(I) = \frac{1}{\left|w^2\right|}\sum_{k:(i,j)\in w_{k}}\left(1+\frac{(I_{i}-\mu_{k})(I_{j}-\mu_{k})}{\sigma_{k}^{2}+\varepsilon}\right)
 \label{eq:coefficient}
 \end{equation}
 where $\left|w\right|$ is the number of pixels in the local square window $w_{k}$, $\mu_{k}$ and $\sigma^{2}$ are the mean and variance of the image patch in $w_{k}$, and $\varepsilon$ is a regularization parameter. In this way, we seamlessly integrate the recognition capacity of DCNNs and the fine-grained localization ability of guided filter. The object masks can further be refined by using the guidance image to fine-tune the output of DCNNs.

 \subsection{High-Level Guided Low-Level (HGL) Module}
In the GF module, the low-level features are often used as the guidance image because they contain more object structures. In fact, the low-level features are too noisy to provide a high-quality guidance. Using the low-level features directly as the guidance image could be less effective for semantic image segmentation. Actually, the high-level features can guide the low-level features on recovering the localization details of objects. To this end, we propose High-level Guided Low-level (HGL) module, which integrates the high-level features to extract more discriminative low-level features:
\begin{equation}
   HGL(X) = HF(X)\circledast LF(X)
   \label{eq:HGL}
 \end{equation}
 where $HF(X)$ and $LF(X)$ are high-level features and low-level features, respectively. $\circledast$ represents the element-wise multiplication.
 The detailed structure of HGL module is shown in Figure~\ref{fig:network}. By using the HGL module, the quality of the guidance image can be improved, which can further boost the performance of semantic image segmentation.

\subsection{Complete Network Architecture}
The ResNet-101 model~\cite{he2016deep} has achieved promising performance on image classification and the Modified Aligned ResNet-101 used in DeepLabv3+~\cite{chen2018encoder} has also demonstrated its strong potential for the task of semantic image segmentation. On top of that, we use Atrous Spatial Pyramid Pooling module (ASPP) proposed in the DeepLabv2~\cite{chen2017deeplab} to capture multi-scale information. With our designed BG, GF and HGL, we propose a Guided Filter Network (GFN) to achieve semantic image segmentation, as illustrated in Figure~\ref{fig:network}.

In this paper, we use the cross-entropy loss function for semantic image segmentation, which is formulated as follows:
 \begin{equation}
   Loss = -\frac{1}{h \times w}\sum_{i=1}^h\sum_{j=1}^w\sum_{c=1}^Cp_{ij}^{c}\log S_{ij}^{c}
 \label{eq:output_loss}
 \end{equation}
 where $C$ is the number of classes and $p_{ij}^{c}$ is the ground truth probability of the $c^{th}$ class at the pixel $(i,j)$. $S_{ij}^{c}$ represents the posterior probability of pixel $(i,j)$ belonging to the $c^{th}$ class. $h$ and $w$ are the height and the width of the image, respectively.

\begin{figure*}[t]
\begin{center}
\includegraphics[width=1.0\linewidth]{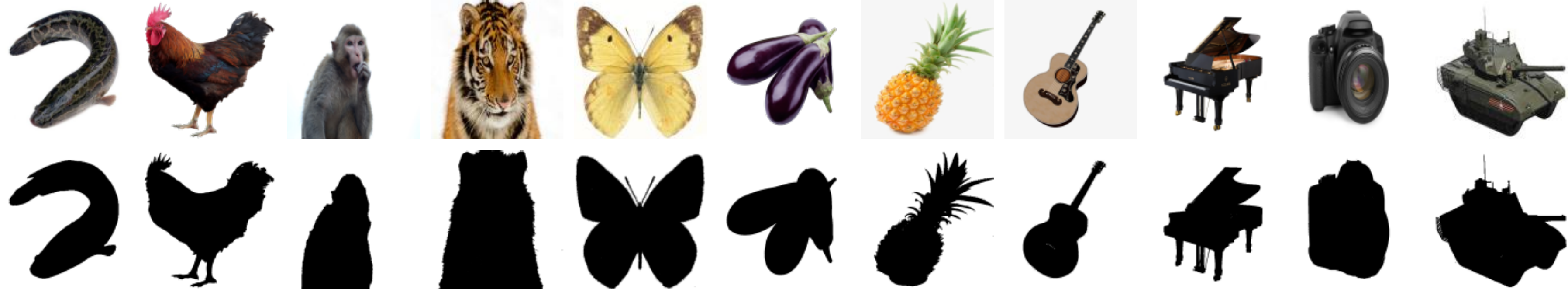}
\end{center}
\vspace*{-0.38cm}
\caption{The Otsu detector can be used to extract the object masks from simple images with homogeneous backgrounds.}
\label{fig:otsu}
\vspace*{-0.28cm}
\end{figure*}

\section{Generation of Pixel-Level Semantic Labellings}

The overall architecture of our proposed method is illustrated in Figure~\ref{fig:framework}. To generate fine-grained object masks automatically, our proposed approach takes three key steps: (a) the GFN is first trained in the source domain, where the pixel-level labellings are available for the training images; (b) the GFN is then used to produce initial coarse object masks with object boundaries for the images in the target domain; (c) such coarse object masks are treated as pseudo labels and they are further leveraged to optimize/refine the GFN iteratively.

\subsection{Pretext Tasks design for Generating Initial Object Masks}

Compared with supervised learning methods that require manual labelling, the proposed approach requires initial object masks which can be automatically generated by transferring the segmentation knowledge learned from the source domain to the target domain. The learning of segmentation knowledge depends on the design of the pretext task, and a good pretext task can effectively improve the quality of the initial object masks. Considering that the goal of self-supervised learning is to generate initialization parameters for downstream tasks by proxy task that do not require human annotations, we design three pretext tasks to learn segmentation knowledge. After the pretext task training finished, the learned parameters are transferred to GFN to generate initial object masks in the target domain.

\subsubsection{A transfer learning approach}
For a given object class of interest, its semantically-similar category is first identified from public datasets, and the training images with pixel-level labellings for its semantically-similar category are used to train its GFN. Such learned GFN is further used to obtain the initial object masks for the images in the target domain. For example, for a given object class of interest ``angraecum-mauritianum'' (fine-grained plant species) in the ``Orchid'' family, we do not have its training images with pixel-level labellings, but we may have sufficient training images that are labelled at the pixel level for its semantically-similar category ``plant'' in public datasets such as PASCAL VOC, Microsoft COCO and BSD. Thus we can leverage the training images for the semantically-similar category ``plant'' (in the source domain) to train the GFN for the given object class ``angraecum-mauritianum'' (in the target domain) as illustrated in the left of Figure~\ref{fig:framework} (a), and the GFN can further be used to obtain the initial masks for the object class ``angraecum-mauritianum''.

\subsubsection{A simple-to-complex approach}
For some object classes of interest, their semantically-similar categories cannot be identified from public datasets, thus our transfer learning approach cannot be used and a simple-to-complex approach is developed to handle this situation. To generate the initial object masks for a given object class of interest, our simple-to-complex approach consists of the following components: (a) some images with simple background are searched from the Internet; (b) Otsu detector~\cite{otsu1979threshold}, as illustrated in Figure~\ref{fig:otsu}, is performed on such simple images with homogeneous backgrounds to generate the masks for the given object class of interest; (c) such object masks are treated as the training images to train GFN to extract initial object masks from the images with more complex backgrounds.

\subsubsection{A Class Activation Maps (CAMs) approach}
One drawback of the simple-to-complex approach is the requirement of manual selection of the images with simple backgrounds. To solve this problem, we use CAMs~\cite{zhou2016learning} to generate the initial object masks. We follow the approach in~\cite{zhou2016learning} to learn CAMs whose architecture is typically a classification network with global average pooling (GAP) followed by a fully-connected layer. We train the CAMs from the images with image-level labellings by using a classification criteria. The learned CAMs is then used to generate the object masks from the images in the target domain, and such object masks are further refined by using dCRF~\cite{krahenbuhl2011efficient}. The segmentation label for a training image can be obtained by selecting the class label associated with the maximum activation score at every pixel in the up-sampled CAMs~\cite{ahn2018learning}.

\subsection{Iterative Learning of GFN for labellings generation}
For a given set of object classes of interest, after the GFN is learned from the training images in the source domain (pretext task), it is then used to obtain the initial masks for the object classes of interest in the target domain. Because of the significant difference between the source domain and the target domain, the GFN may not be able to obtain the initial object masks accurately in the target domain. One reasonable solution is to fine-tune the GFN iteratively in the target domain. For this purpose, we treat the refined object masks as a pseudo label to optimize/refine GFN iteratively. Note that noisy labellings may lead to a drift in semantic image segmentation when they are used to supervise the training of GFN directly. Thus we define the reliability of a labelling (generated by our GFN) based on the area ratio, e.g., the proportion of the area between the automatically-labelled region and the whole image. The area ratio can be formulated as:
\begin{equation}
S_{ratio} = \frac{S_{annotation}}{S_{image}}
\end{equation}
where $S_{annotation}$ represents the sum of pixels of the automatically-labelled region, and $S_{image}$ is the sum of pixels of the whole image. Only a subset of the pixels in the automatically-labelled region within a ratio between 0.1 and 0.9 are selected for network training, and this parameter can appropriately be adjusted according to the ratio of the object's area in the image. We choose such parameters because the objects in our image set are neither too big nor too small. By continuously leveraging such training images with automatically-generated pixel-level labellings to optimize/refine our GFN in the target domain, the generated object masks may become more accurate gradually. After several iterations, we can finally obtain the fine-grained object masks with detailed pixel-level structures/boundaries in the target domain.

\section{Experimental Results and Analysis}
This section describes our experimental results and our analysis for algorithm evaluation over multiple image datasets.

\subsection{Training details}
To train the GFN, a mini-batch size of 8 images is used. The network parameters are initialized by the ResNet-101 model~\cite{he2016deep} which is pre-trained on ImageNet~\cite{deng2009imagenet}. The initial learning rate is set as 0.007 and divided by 10 in every 5 epochs. The weight decay and the momentum are set to 0.0002 and 0.9, respectively. In addition, in the Guided Filter (GF) Module, the values of $r$ and $\varepsilon$ are first determined by grid search on the validation set. We then use the same parameters to train GFN.

\subsection{Generating pixel-level labellings over ``Orchid'', FGVC Aircraft, CUB-200-2011 and Stanford Cars Image Sets}

\begin{itemize}
\item \textbf{PASCAL VOC 2012:} The PASCAL VOC 2012 segmentation benchmark~\cite{everingham2015pascal} contains 20 foreground object classes and one background class. It has 1,464, 1,449, and 1,456 images labelled at the pixel level for network training, validation, and testing, respectively. The dataset is augmented by the extra labellings provided by~\cite{hariharan2011semantic}, resulting in 10,582 training images.

\item \textbf{``Orchid'' Plant Image Set:} We have crawled large-scale plant images which include 252 plant species (object classes) in the ``Orchid'' family, where 149,500 plant images are used to train the base network and other 500 images for testing.

\item \textbf{FGVC Aircraft:} FGVC Aircraft dataset~\cite{maji2013fine} contains 10,000 images for 100 fine-grained aircraft classes. We divide 10,000 images into 9,000 training images and 1,000 test images.

\item \textbf{CUB-200-2011:} The CUB-200-2011 dataset~\cite{xiao2015application} contains 200 fine-grained bird species. It has 11,788 images, of which 11,288 images for training, 500 images for testing.

\item \textbf{Stanford Cars:} The Stanford Cars dataset~\cite{krause20133d} contains 16,185 images for 196 fine-grained classes. We divide 16,185 images into 15,685 training images and 500 test images.
\end{itemize}

Note that the test images of the last four datasets are manually labelled at the pixel level. We use our proposed approach to generate pixel-level labellings, where the initial object masks are generated by our transfer learning approach.

It is worth noting that our transfer learning approach can generate initial object masks directly, where PASCAL VOC 2012 is treated as the source domain and our ``Orchid'' plant, FGVC Aircraft, CUB-200-2011, and Stanford Cars image sets are treated as the target domain. The reason is that the categories of ``potted plant'', ``aeroplane'', ``bird'', and ``car'' in the PASCAL VOC 2012 is semantically-similar with all 252 plant species in our ``Orchid'' plant image set, 100 fine-grained ``aircraft'' classes in the FGVC Aircraft image set, 200 fine-grained ``bird'' classes in the CUB-200-2011 image set, and 196 fine-grained ``car'' classes in the Stanford Cars image set, respectively. Thus the knowledge learned from the PASCAL VOC 2012 dataset is transferable to our target datasets.

\subsection{Generating pixel-level labellings over Branch Image Set}

Our Branch image set contains 2,134 images, including 500 complex images and 1,634 simple images. Images are further divided into the training set and the test set, which have 2,034 and 100 images respectively. The 100 test images are manually labelled at the pixel level as the ground truth.

For the ``branch'' class, its semantically-similar category cannot be identified from public datasets, we use our proposed approach to generate pixel-level labellings, where the initial object masks are generated by our simple-to-complex approach.

We have also evaluated the impacts of the number of simple images with pixel-level labellings on the performance of our proposed method. As illustrated in Table~\ref{tab:branch_data_result}, one can easily observe that the performance of our proposed method can be improved when more simple images are incorporated for network training.

\begin{table}
\small
\caption{Comparison on the performance of our GFN networks when different numbers of images are used.}
\begin{center}
\begin{tabular}{|l|c|c|c|c|}
\hline
Number of Simple Images & 205 & 409 & 817 & 1634  \\
\hline\hline
IoU ($\%$) & 58.5 & 60.5 & 64.8 & 70.1 \\
\hline
\end{tabular}
\end{center}
\label{tab:branch_data_result}
\end{table}

\subsection{Generating pixel-level labellings over Pascal Voc 2012}

For the PASCAL VOC 2012 dataset, we use our proposed approach to generate pixel-level labellings, where the initial object masks are generated by the CAMs approach~\cite{zhou2016learning}.

\begin{figure}[t]
\begin{center}
\includegraphics[width=0.85\linewidth]{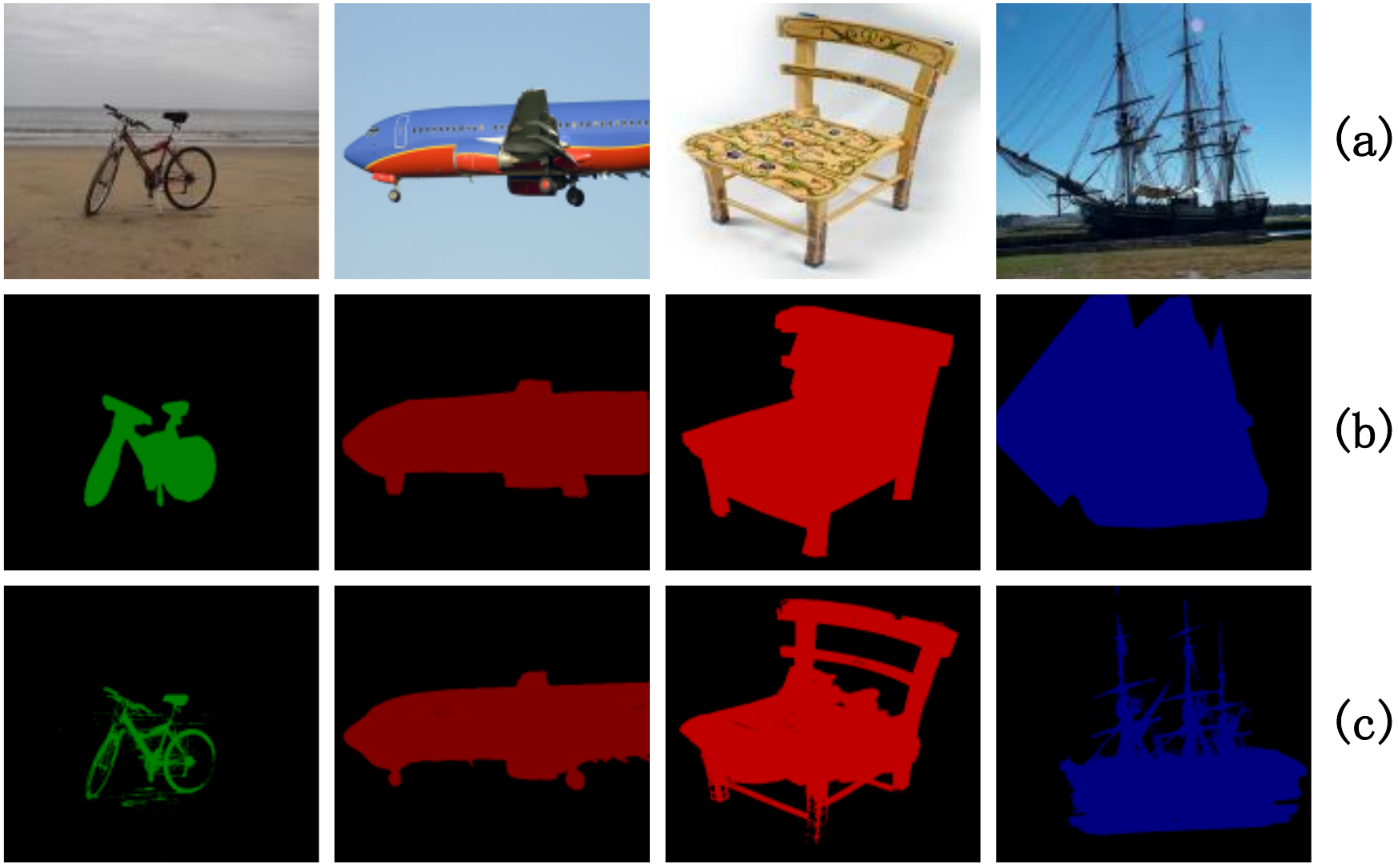}
\end{center}
\vspace*{-0.48cm}
  \caption{Automatically-generated labellings and Ground Truth on examples selected from PASCAL VOC 2012 dataset. We show the examples where automatically-generated labellings are more accurate than ground truth. (a) original image; (b) ground truth; (c) the final fine-grained object masks.}
\label{fig:humans_vs_generate}
\vspace*{-0.38cm}
\end{figure}

Figure~\ref{fig:humans_vs_generate} provides some of our segmentation results on the PASCAL VOC 2012 training set. From these experimental results, we observe that the automatically-generated labellings have precise segmentation masks even some labellings are better than ground truth. We attribute this to our proposed Guided Filter Network (GFN) that can transfer the structure of the object class of interest into the segmentation result to reason refined masks. Meanwhile, this indicates that manually labelling large-scale training images at the pixel level is very difficult to provide consistent labelling quality.

\subsection{Select the number of iterations}
We use mIoU to verify the effectiveness of our proposed method and calculate its mIoU after every iteration until the mIoU no longer grows. The model with the best mIoU is then selected to generate the final pixel-level labellings. For the iterative training method, we start from the segmentation results for the first iteration, in later iterations, the segmentation results for the previous iteration are employed to train the segmentation network for the next iteration.

\begin{figure}[t]
\begin{center}
\includegraphics[width=0.85\linewidth]{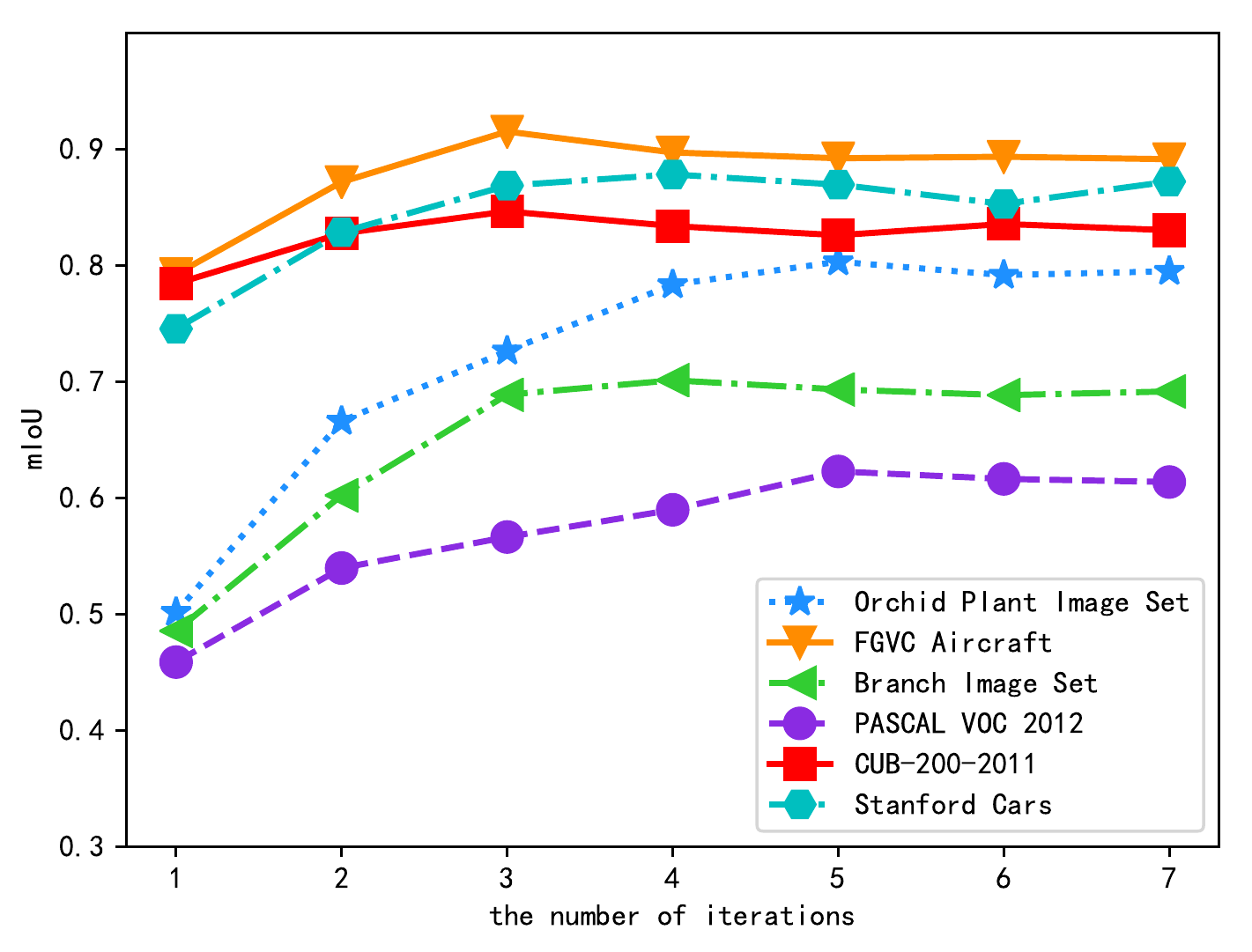}
\end{center}
\vspace*{-0.38cm}
   \caption{mIoU of our proposed method is evaluated at every iteration in the ``Orchid'' Plant image set, FGVC Aircraft image set, CUB-200-2011 image set, Stanford Cars image set, Branch image set, and the validation set of PASCAL VOC 2012, respectively.}
\label{fig:iterations}
\vspace*{-0.28cm}
\end{figure}

\begin{figure}[t]
\begin{center}
\includegraphics[width=0.85\linewidth]{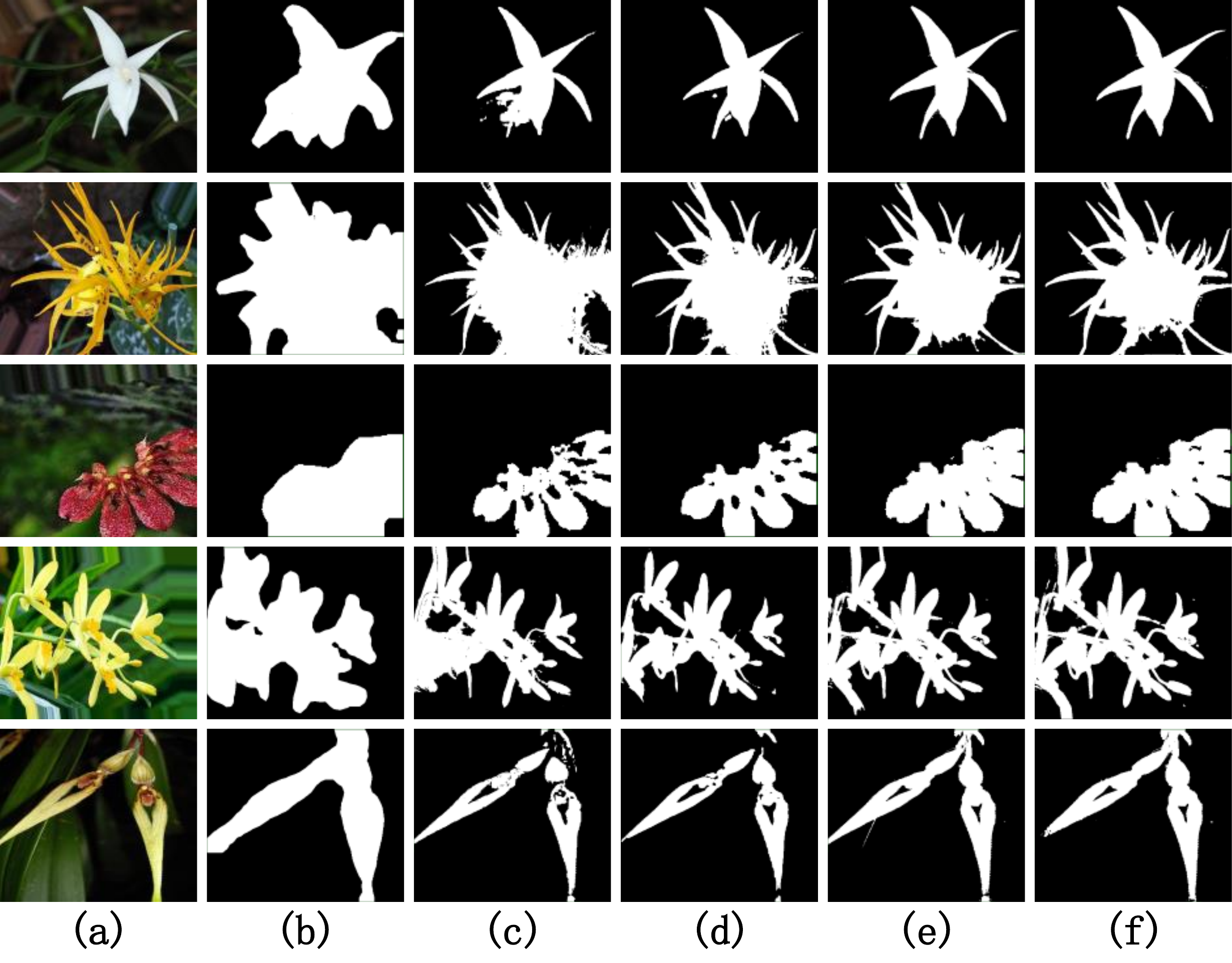}
\end{center}
\vspace*{-0.38cm}
   \caption{The object masks extracted by our proposed approach on ``Orchid'' Plant image set: (a) original images; (b) initial object masks obtained by our transfer learning approach; (c)-(e) refined object masks at different iterations; (f) the final fine-grained object masks.}
\label{fig:plant_complex}
\vspace*{-0.28cm}
\end{figure}

\begin{figure}[t]
\begin{center}
\includegraphics[width=0.85\linewidth]{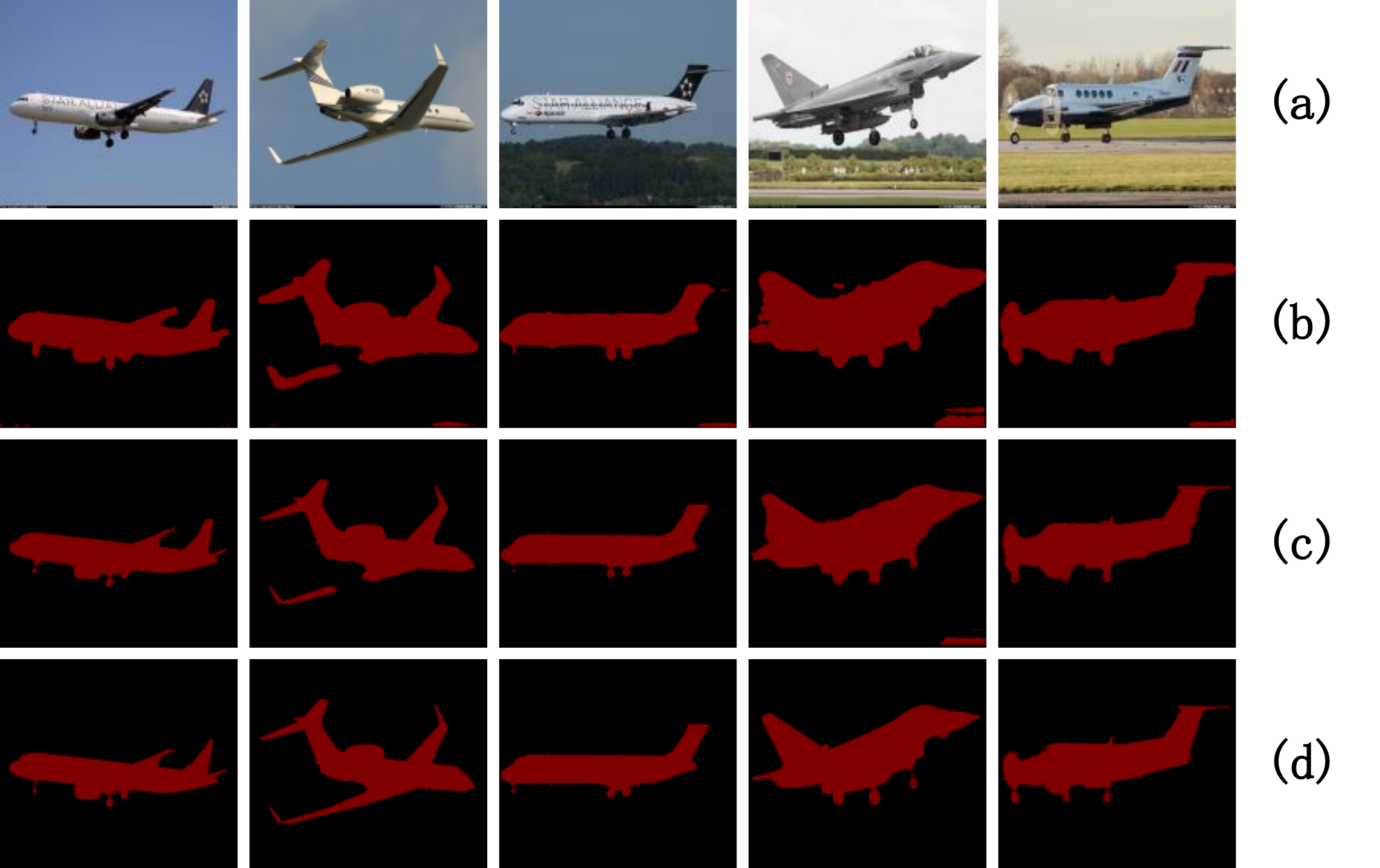}
\end{center}
\vspace*{-0.38cm}
  \caption{The object masks extracted by our proposed approach on FGVC Aircraft image set: (a) original images; (b) initial object masks obtained by our transfer learning approach; (c) refined object masks at different iterations; (d) the final fine-grained object masks.}
\label{fig:aircraft_process}
\vspace*{-0.28cm}
\end{figure}

\begin{figure}[t]
\begin{center}
\includegraphics[width=0.85\linewidth]{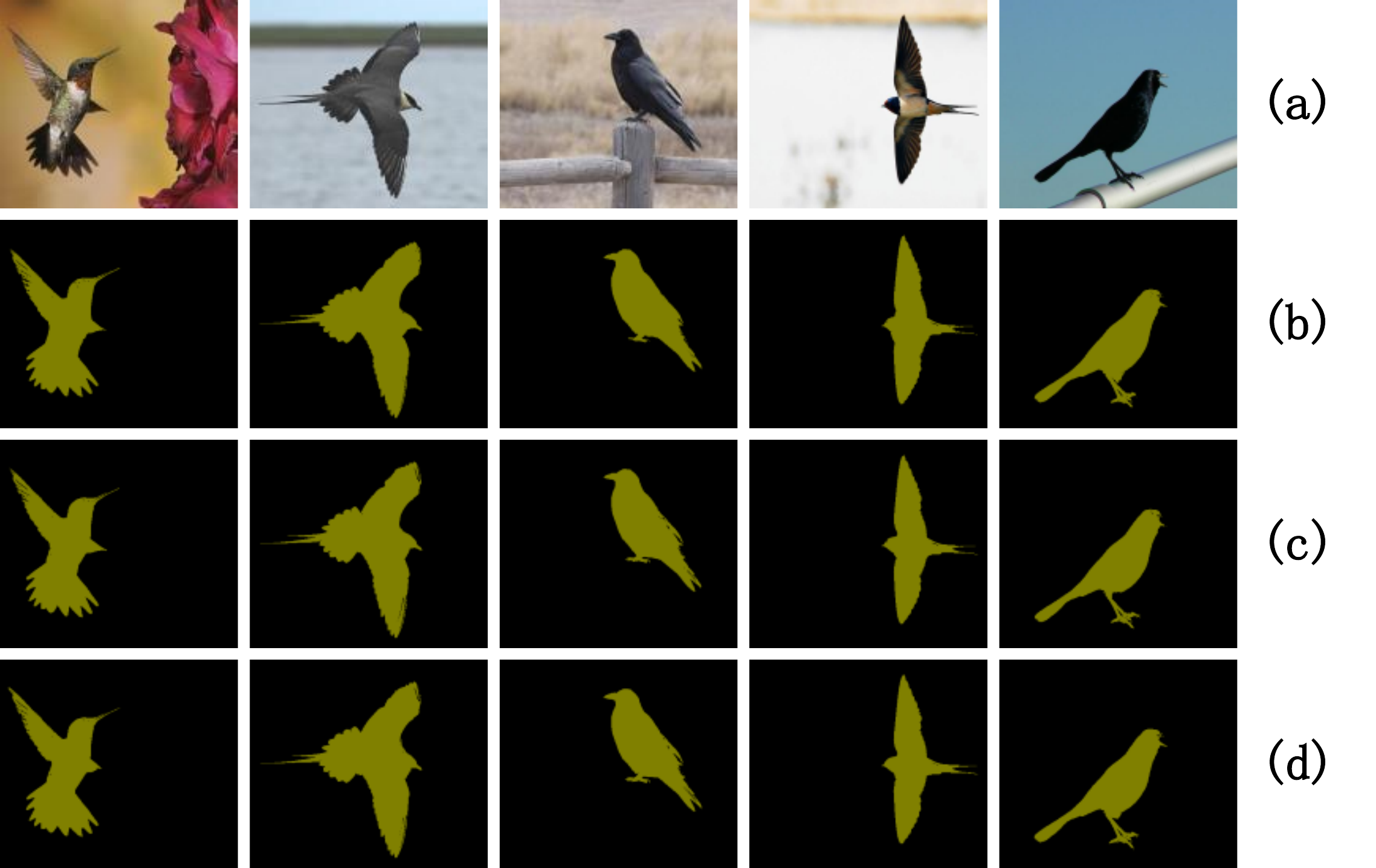}
\end{center}
\vspace*{-0.38cm}
  \caption{The object masks extracted by our proposed approach on CUB-200-2011 image set: (a) original images; (b) initial object masks obtained by our transfer learning approach; (c) refined object masks at different iterations; (d) the final fine-grained object masks.}
\label{fig:Bird_process}
\vspace*{-0.28cm}
\end{figure}

\begin{figure}[t]
\begin{center}
\includegraphics[width=0.85\linewidth]{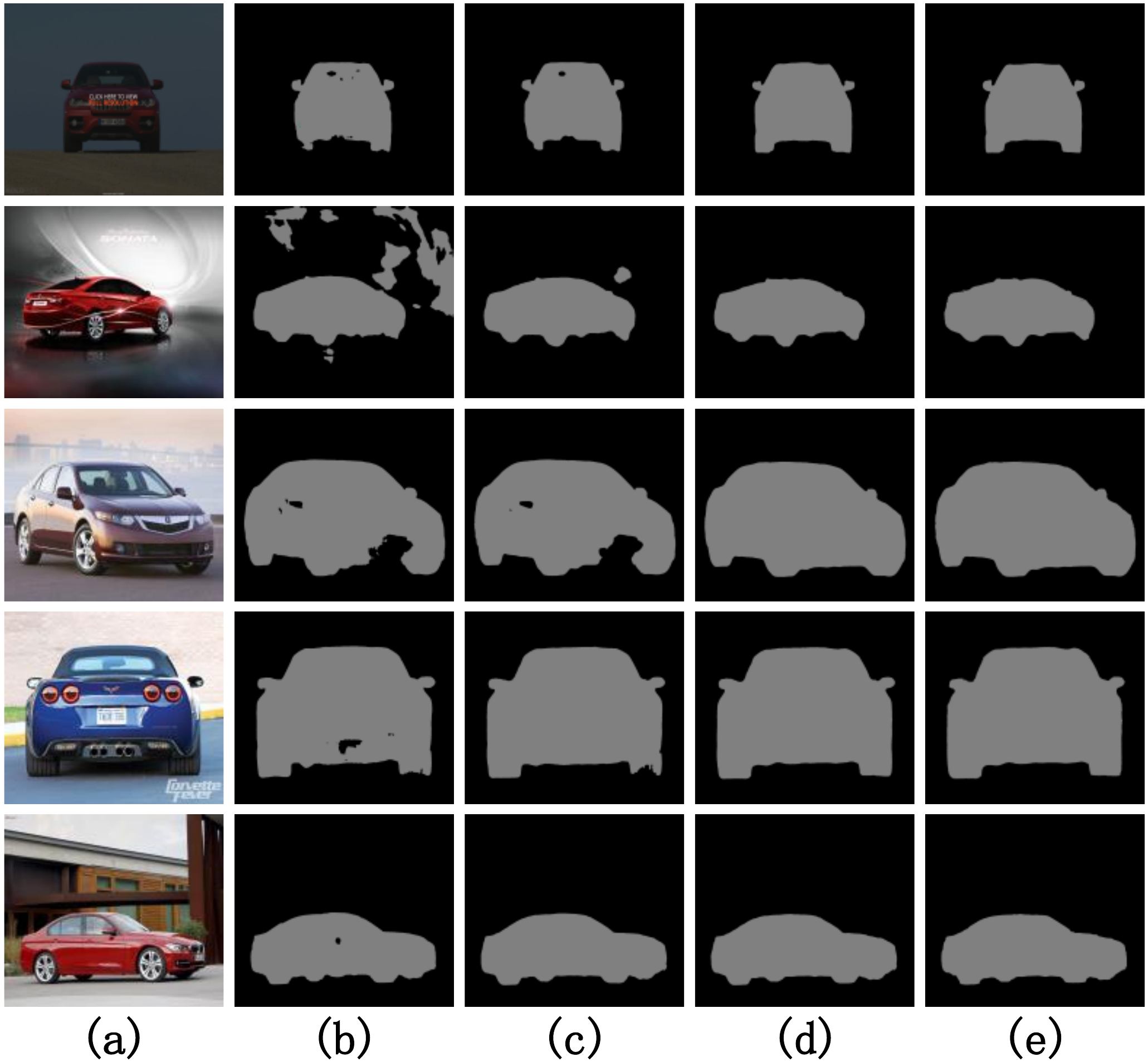}
\end{center}
\vspace*{-0.38cm}
  \caption{The object masks extracted by our proposed approach on Stanford Cars image set: (a) original images; (b) initial object masks obtained by our transfer learning approach; (c)-(d) refined object masks at different iterations; (e) the final fine-grained object masks.}
\label{fig:Car_process}
\vspace*{-0.28cm}
\end{figure}

\begin{figure}[t]
\begin{center}
\includegraphics[width=0.85\linewidth]{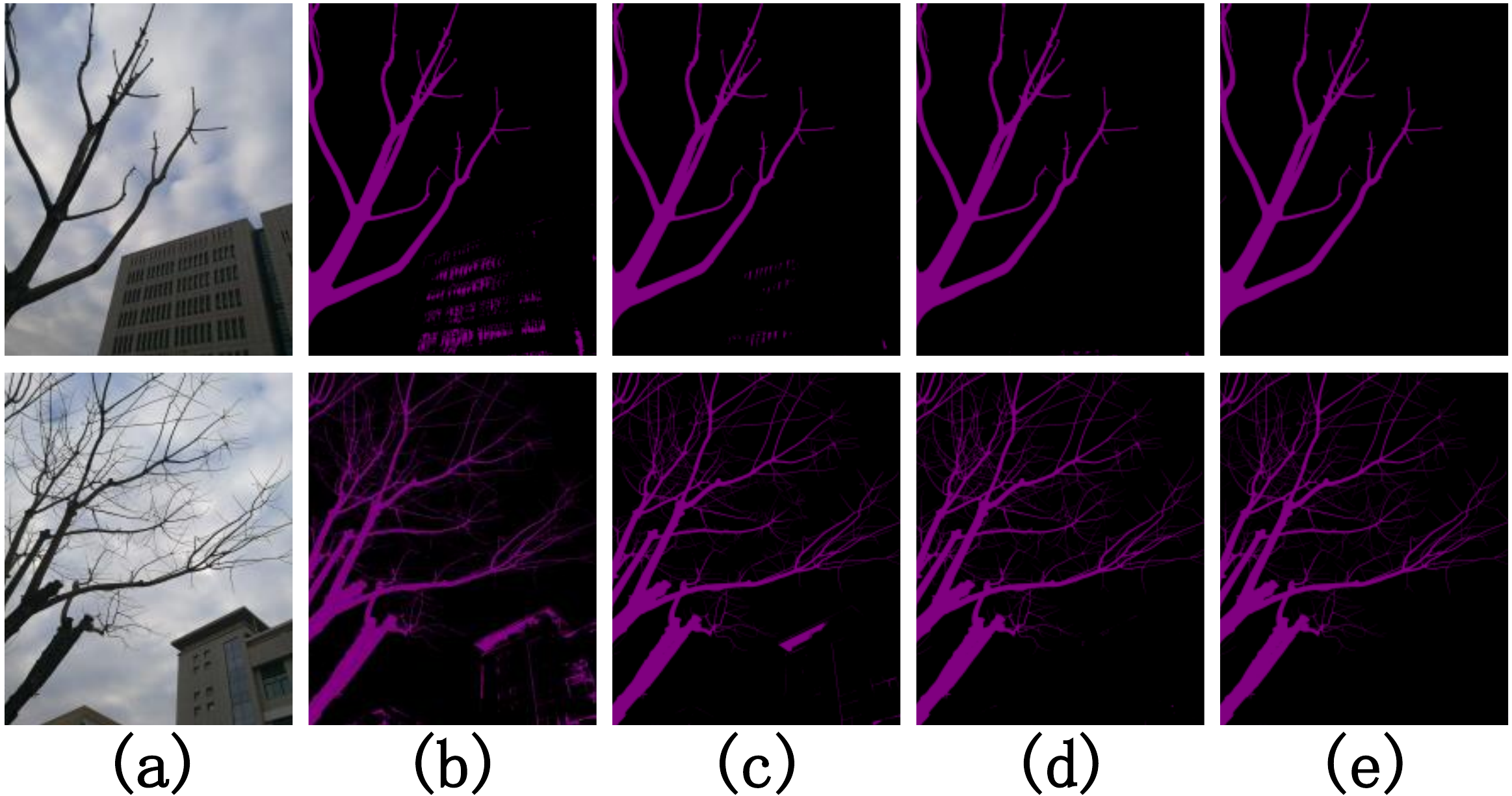}
\end{center}
\vspace*{-0.38cm}
  \caption{The object masks extracted by our proposed approach on Branch image set: (a) original images; (b) initial object masks obtained by our simple-to-complex approach; (c)-(d) refined object masks at different iterations; (e) the final fine-grained object masks.}
\label{fig:brand_process}
\vspace*{-0.28cm}
\end{figure}

Figure~\ref{fig:iterations} shows the performance of the recursive refinement. The performance of all these networks improves as the training round increases at first, and then saturates after few training rounds. Some examples for each step of our proposed approach are shown in Figure~\ref{fig:plant_complex}, Figure~\ref{fig:aircraft_process}, Figure~\ref{fig:Bird_process}, Figure~\ref{fig:Car_process} and Figure~\ref{fig:brand_process}.

\subsection{Evaluation of automatically-generated labellings}

\begin{figure*}[t]
\begin{center}
\includegraphics[width=0.9\linewidth]{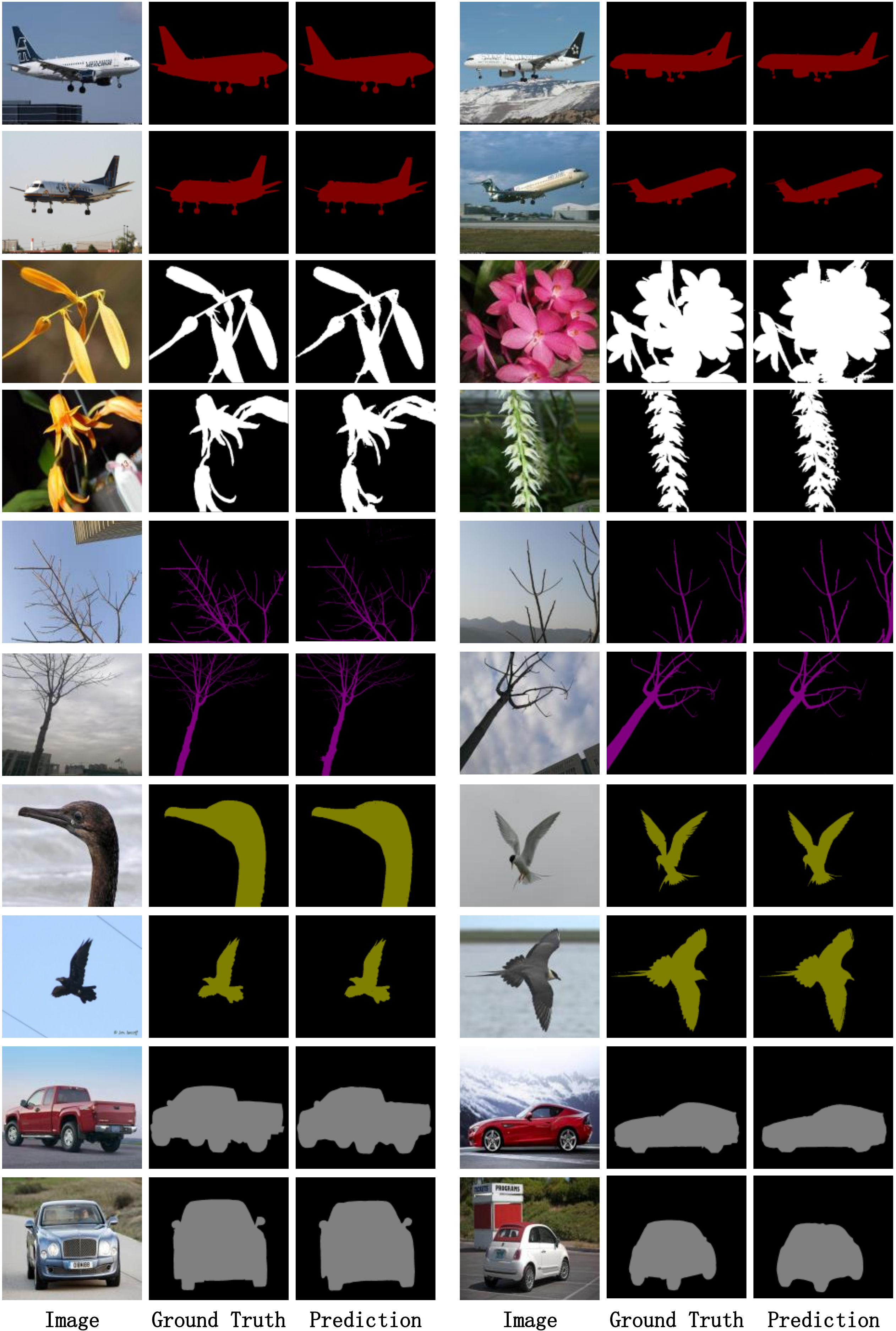}
\end{center}
\vspace*{-0.38cm}
  \caption{The results obtained by our proposed approach on ``Orchid'' Plant, FGVC Aircraft, CUB-200-2011, Stanford Cars, and Branch image sets.}
\label{fig:combine_supplementary_material}
\vspace*{-0.28cm}
\end{figure*}

\begin{figure*}[t]
\begin{center}
\includegraphics[width=0.9\linewidth]{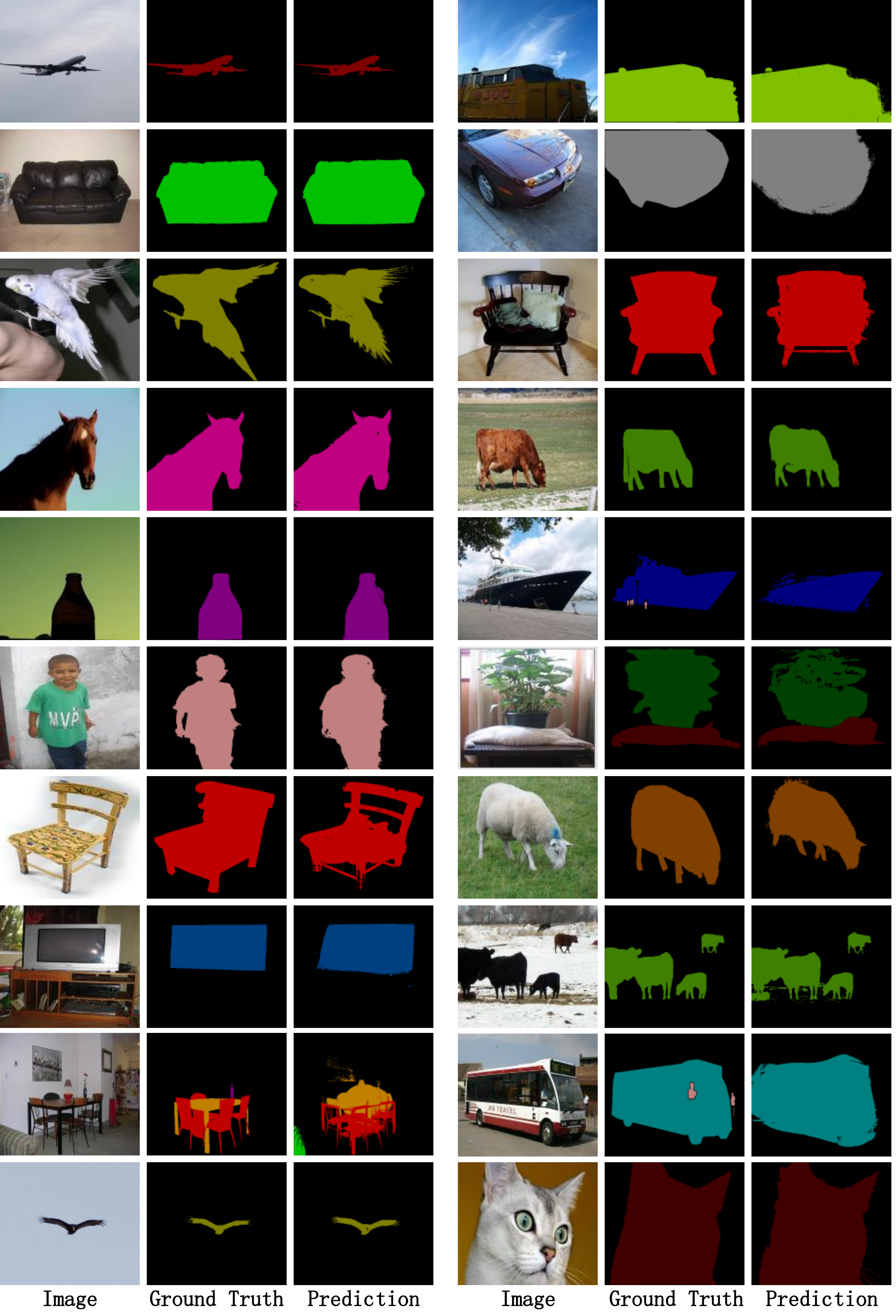}
\end{center}
\vspace*{-0.38cm}
  \caption{The segmentation results obtained by our proposed approach on PASCAL VOC 2012 training set.}
\label{fig:pascal_voc_supplementary_material}
\vspace*{-0.28cm}
\end{figure*}

We use mIoU as our main evaluation metric. Table~\ref{tab:branch_tree_plant_test} summarizes the evaluation results on the mIoU over ``Orchid'' Plant, CUB-200-2011, Stanford Cars, FGVC Aircraft, Branch, and PASCAL VOC 2012 image sets. One can easily observe that our proposed approach can achieve good performance on generating pixel-level labellings (e.g., fine-grained object masks) with acceptable mIoU. Figure~\ref{fig:combine_supplementary_material} and Figure~\ref{fig:pascal_voc_supplementary_material} describe several fine-grained object masks with detailed pixel-level structures/boundaries on ``Orchid'' Plant, FGVC Aircraft, CUB-200-2011, Stanford Cars, Branch, and PASCAL VOC image sets. From these experimental results, one can easily observe that our proposed approach can automatically generate fine-grained object masks, and the annotation quality is very close to that of manually-labelled ones.

\begin{table}
\caption{The mIoU ($\%$) over six image sets by our proposed approach.}
\begin{center}
\begin{tabular}{|l|c|c|c|}
\hline
Dataset  & mIoU ($\%$) & Dataset  & mIoU ($\%$)\\
\hline\hline
``Orchid'' Plant & 80.3 & CUB-200-2011 & 84.6 \\
\hline
Stanford & 87.8 & FGVC Aircraft & 91.5 \\
\hline
Branch & 70.1 & PASCAL VOC 2012 & 62.3 \\
\hline
\end{tabular}
\end{center}
\label{tab:branch_tree_plant_test}
\end{table}

There is an experiment that could be carried out to shed some light in terms of performance. We provide the comparison results when the manually-labelled images and the automatically-labelled images (whose annotations at the pixel level are automatically generated by our proposed method) are employed for network training. Thus, we have evaluated two approaches on their mIoU when: (a) manually-labelled images are used, where we choose 800 manually-labelled images in the FGVC Aircraft image set as the training images and 200 as the test images; (b) automatically-labelled images are used, where the pixel-level labellings for these 800 training images are automatically generated by our proposed method. We use our GFN as the baseline model, where the parameters are initialized by the ResNet-101 model which is pre-trained on ImageNet. Table~\ref{tab:manually-generated-miou} summarizes the evaluation results on mIoU, one can easily observe that our proposed approach (i.e., leveraging automatically-labelled images for network training) can achieve very competitive performance.

\begin{table}
\caption{The comparison results on mIoU ($\%$) when manually-labelled images and the automatically-labelled images are used for network training.}
\begin{center}
\begin{tabular}{|l|c|}
\hline
Training Images  & mIoU ($\%$)\\
\hline\hline
Manually-Labelled Ones & 92.3 \\
\hline
Automatically-Labelled Ones  & 88.3 \\
\hline
\end{tabular}
\end{center}
\label{tab:manually-generated-miou}
\end{table}

\subsection{Comparison with Weakly-Supervised Methods}
\subsubsection{PASCAL VOC 2012}
Our proposed approach share some similar principles with weakly-supervised methods, e.g. fine-tuning image-level coarse labellings into pixel-level ones, and our proposed method is compared against numerous up-to-date weakly supervised approaches. The segmentation results on PASCAL VOC 2012 validation set are quantified and compared as illustrated in Table~\ref{tab:weakly_supervised_deeplabv3_plus}. We achieve the mIoU score of 62.3 on PASCAL VOC 2012 validation set, which outperforms most the previous approaches using the same supervision settings. In order to accomplish a fair comparison, we use VGG16~\cite{simonyan2014very} as an encoder. We achieve the mIoU score of 59.4 on PASCAL VOC 2012 validation set, which outperforms other methods in weakly-supervised semantic image segmentation. Moreover, we want to emphasize that the proposed method only use image-level labelling as supervision without the information of external sources.

From Table~\ref{tab:weakly_supervised_deeplabv3_plus}, we observe that the results of ~\cite{zeng2019joint} and ~\cite{wang2020self} on the PASCAL VOC 2012 validation set are slightly better than our method. One possible reason is that the PASCAL VOC 2012 containing very few images per class, and our proposed approach can improve the segmentation performance by incorporating more images for network training (see Table~\ref{tab:branch_data_result}). When large-scale images are used for training, these methods may be far inferior to our method. The specific experimental verification is shown in the Section~\ref{section:comparison_with_weakly-supervised_methods} of this paper.

\begin{table}
\small
\caption{Strategies and results on PASCAL VOC 2012 validation set.}
\begin{center}
\begin{tabular}{|l|c|}
\hline
 Methods & mIoU(\%)  \\
\hline\hline
\textbf{Image-level labels with external source} &   \\
STC~\cite{wei2016stc} & 49.3     \\
Co-segmentation~\cite{shen2017weakly} & 56.4     \\
Webly-supervised~\cite{kolesnikov2016seed} & 53.4    \\
Crawled-Video~\cite{hong2017weakly} & 58.1     \\
AF-MCG~\cite{qi2016augmented} & 54.3     \\
Joon \emph{et al.} ~\cite{oh2017exploiting} & 55.7     \\
DCSP-VGG16~\cite{chaudhry2017discovering} & 58.6    \\
DCSP-ResNet-101~\cite{chaudhry2017discovering} & 60.8     \\
Mining-pixels~\cite{hou2016mining} & 58.7     \\
\hline\hline
\textbf{Image-level labels w/o external source} &    \\
MIL-FCN~\cite{pathak2014fully} & 25.7     \\
EM-Adapt~\cite{papandreou2015weakly} & 38.2    \\
BFBP~\cite{saleh2016built} & 46.6     \\
DCSM~\cite{shimoda2016distinct} & 44.1     \\
SEC~\cite{kolesnikov2016seed} & 50.7     \\
AF-SS~\cite{qi2016augmented} & 52.6     \\
SPN~\cite{kwak2017weakly} & 50.2 \\
Two-phase~\cite{kim2017two} & 53.1     \\
Anirban \emph{et al.}~\cite{roy2017combining} & 52.8     \\
AdvErasing~\cite{wei2017object} & 55.0     \\
DCNA-VGG16~\cite{zhang2019decoupled}  & 55.4     \\
MCOF~\cite{wang2018weakly} & 60.3 \\
MDC~\cite{wei2018revisiting} & 60.4 \\
DSRG~\cite{huang2018weakly} & 61.4 \\
AffinityNet~\cite{ahn2018learning} & 61.7 \\
Zeng \emph{et al.}~\cite{zeng2019joint} & 64.3 \\
SEAM~\cite{wang2020self} & 64.5 \\
\textbf{Ours-VGG16}  & \textbf{59.4}     \\
\textbf{Ours-ResNet-101}  & \textbf{62.3}     \\
\hline
\end{tabular}
\end{center}
\label{tab:weakly_supervised_deeplabv3_plus}
\end{table}

\subsubsection{Branch+Aircraft+Plant}
\label{section:comparison_with_weakly-supervised_methods}
We have compared our proposed approach with some weakly-supervised methods~\cite{papandreou2015weakly,kolesnikov2016seed,ahn2018learning,wang2020self} on generating fine-grained object masks on the Branch+Aircraft+Plant image set. As illustrated in Table~\ref{tab:comp2_weakly}, one can easily observe that our proposed approach can achieve better performance on generating fine-grained object masks than the state-of-the-art weakly-supervised methods. Interestingly, one can also find that our proposed approach actually gives slightly better performance on PASCAL VOC 2012 validation set, and there is a significant improvement on Branch+Aircraft+Plant image set. This is due to the PASCAL VOC 2012 containing very few images per class, and our proposed approach can improve the segmentation performance by incorporating more images for network training (see Table~\ref{tab:branch_data_result}).

Unlike these weakly-supervised methods which directly leverage coarsely-labelled images for network training, our proposed approach first fine-tune image-level coarse labellings into pixel-level fine-grained ones, and then leverage such training images with pixel-level labellings to learn more discriminative networks for semantic image segmentation.

\begin{itemize}
\item \textbf{Branch+Aircraft+Plant:} We merged the Branch Image set, FGVC Aircraft dataset and ``Orchid'' Plant Image set, 25,822 images for training (2,034 images from Branch image set, 9,000 images from FGVC Aircraft dataset, 14,788 images from ``Orchid'' plant image set) and 1,600 for testing (of which 100 branch images are from Branch image set, 1,000 images from FGVC Aircraft dataset, 500 images from ``Orchid'' plant image set). To avoid data imbalance, we only select 14,788 ``Orchid'' plant images for network training.
\end{itemize}

\begin{table}
\small
\caption{The comparison results on various object classes and mIoU ($\%$).}
\begin{center}
\begin{tabular}{|l|c|c|c|c|c|}
\hline
 Methods & bkg & aircraft & plant & branch & mIoU \\
\hline\hline
EM-Adapt~\cite{papandreou2015weakly} & 33.6 & 28.4 & 35.8 & 11.2 & 27.3 \\
\hline
SEC~\cite{kolesnikov2016seed} & 62.0 & 61.0 & 65.7 & 28.3 & 54.3 \\
\hline
AffinityNet~\cite{ahn2018learning} & 84.3 & 58.0 & 51.4 & 20.9 & 53.6 \\
\hline
SEAM~\cite{wang2020self} & 87.7 & 57.8 & 50.0 & 19.4 & 53.7 \\
\hline
\textbf{Ours-VGG16} & \textbf{92.7} & \textbf{88.3} & \textbf{74.3} & \textbf{65.4} & \textbf{80.2} \\
\hline
\textbf{Ours-ResNet-101} & \textbf{94.8} & \textbf{91.1} & \textbf{78.5} & \textbf{68.5} & \textbf{83.2} \\
\hline
\end{tabular}
\end{center}
\label{tab:comp2_weakly}
\end{table}

\subsection{Ablation study}

In this subsection, the effectiveness of our proposed GFN is firstly evaluated over the PASCAL VOC 2012 segmentation benchmark~\cite{everingham2015pascal} against numerous state-of-the-art techniques for semantic image segmentation, and we adopt the mean intersection-over-union (mIoU) as the evaluation metric on segmentation accuracy.

\subsubsection{Ablation study on GF Module}
We use DeepLabv2 as our base network, and the baseline comparison is used to evaluate the effectiveness of our proposed GF. As shown in Table~\ref{tab:stronger_supervised_deeplabv2}, our proposed GF improves the segmentation performance from 71.79$\%$ to 73.82$\%$, which indicates the effects of our proposed GF module.

\subsubsection{Ablation study on HGL Module}
The low-level features are often used as the guidance image because they contain the object structures. However, the low-level features are too noisy to provide a high-quality guidance. Therefore, we integrate the HGL into the GF. As presented in Table~\ref{tab:stronger_supervised_deeplabv2}, one can easily observe that integrating HGL into GF can boost the segmentation performance by a margin of 0.98$\%$ (from 73.82$\%$ to 74.80$\%$).

\subsubsection{Ablation study on BG Module}
By further integrating the BG module into the HGL module, the segmentation performance is advanced from 74.80$\%$ to 75.19$\%$ as shown in Table~\ref{tab:stronger_supervised_deeplabv2}. The BG module facilitates the feature fusion by introducing more semantic information into the low-level features and embedding more spatial information into the high-level features. Interestingly, one can easily observe that BG actually only provides a boost of 0.39$\%$ on the segmentation performance, which is very limited. One possible reason is that part of the spatial information is already encoded by the combination of GF, HGL and ASPP. To enable better assessment of the contribution of each module, three proposed modules are evaluated separately. As shown in Table~\ref{tab:stronger_supervised_deeplabv2}, three proposed modules can effectively improve the segmentation performance from different aspects. In addition, Wu \emph{et al.}~\cite{Wu_2018_CVPR} proposed an end-to-end trainable guided filter module and applied it to semantic image segmentation. Compared with~\cite{Wu_2018_CVPR}, our method achieves better performance on the PASCAL VOC 2012 dataset (75.19\% vs. 73.58\%~\cite{Wu_2018_CVPR}). The reason is that directly using the low-level features as the guidance image may introduce more noise information.

\begin{table}
\small
\caption{Ablation studies on three components of our proposed method, DeepLabv2 is the baseline model.}
\begin{center}
\begin{tabular}{|l|c|c|c|c|c|}
\hline
 Index & Baseline & GF & HGL & BG & mIoU(\%) \\
\hline\hline
1 & \checkmark &  &  &  & 71.79 \\
\hline
2 & \checkmark & \checkmark &  &  & 73.82 \\
\hline
3 & \checkmark &  & \checkmark &  & 72.79 \\
\hline
4 & \checkmark &  &  & \checkmark & 74.49 \\
\hline
5 & \checkmark & \checkmark & \checkmark &  & 74.80 \\
\hline
6 & \checkmark & \checkmark & \checkmark & \checkmark  & 75.19 \\
\hline
\end{tabular}
\end{center}
\label{tab:stronger_supervised_deeplabv2}
\end{table}

It should be emphasized that the three proposed modules have different contributions for different tasks. Based on the results over PASCAL VOC 2012 in Table~\ref{tab:stronger_supervised_deeplabv2}, the BG module is the major source of improvement for fully-supervised learning. As presented in Table~\ref{tab:weakly_supervised_ablation_study}, three proposed modules can effectively improve the segmentation performance over ``Orchid'' Plant image set from different aspects. Interestingly, one can easily observe that the GF module is the major source of improvement for weakly-supervised learning. The reason is that the proposed GF module can transfer the structure of the original image to the filter output, so as to refine the object boundaries and improve the segmentation quality.

\begin{table}
\small
\caption{Ablation studies on three components of our proposed method. Dataset: ``Orchid'' Plant image set. Task: weakly-supervised semantic segmentation. DeepLabv2 is the baseline model.}
\begin{center}
\begin{tabular}{|l|c|c|c|c|c|}
\hline
 Index & Baseline & GF & HGL & BG & mIoU(\%) \\
\hline\hline
1 & \checkmark &  &  &  & 67.43 \\
\hline
2 & \checkmark & \checkmark &  &  & 76.29 \\
\hline
3 & \checkmark &  & \checkmark &  & 68.14 \\
\hline
4 & \checkmark &  &  & \checkmark & 69.08 \\
\hline
5 & \checkmark & \checkmark & \checkmark & \checkmark  & 80.30 \\
\hline
\end{tabular}
\end{center}
\label{tab:weakly_supervised_ablation_study}
\end{table}

\subsubsection{Replacing GFN for labellings generation} The GFN can be seen as a segmentation network which can be used for both fully- and weakly-supervised semantic segmentation. The GFN plays a bigger role in weakly-supervised semantic segmentation than in fully-supervised semantic segmentation. The main reason is that fully-supervised learning has been able to obtain good object boundaries, so GFN only further optimized the object details. Since weakly-supervised learning has no effective supervised information, it is difficult to recover the structures/boundaries of the object. The proposed GFN can transfer the structure of the original image to the filter output, so as to refine the object boundaries and improve the segmentation quality. It would be useful to use a common network (e.g., DeepLabv3+) instead of GFN to highlight the true contribution of GFN to generate pixel-level labellings. As shown in Table ~\ref{tab:replacing_gfn}, our approach is superior to DeepLabv3+ for generating pixel-level labellings.

\begin{table}
\caption{The comparison results on mIoU ($\%$) when DeepLabv3+ replacing GFN for weakly-supervised semantic segmentation.}
\begin{center}
\begin{tabular}{|l|c|c|}
\hline
Methods & ``Orchid'' Plant & FGVC Aircraft \\
\hline\hline
DeepLabv3+ & 69.72 & 85.47 \\
\hline
GFN  & 80.30 & 91.50 \\
\hline
\end{tabular}
\end{center}
\label{tab:replacing_gfn}
\end{table}

\section{Conclusions}
A novel approach is proposed in this paper to automatically generate pixel-level object labellings (e.g., fine-grained object masks). Specifically, we first train the GFN in the source domain. Such GFN is then used to identify coarse object masks from the images in the target domain. Such coarse object masks are treated as pseudo labels and they are further leveraged to optimize/refine the GFN iteratively in the target domain. Our experiments on six image sets have demonstrated that our proposed approach can generate fine-grained object masks with detailed pixel-level structures/boundaries, whose quality is very comparable to the manually-labelled ones. In addition, our proposed method can achieve better performance on semantic image segmentation than most existing weakly-supervised approaches.

\ifCLASSOPTIONcaptionsoff
  \newpage
\fi



%



\bibliographystyle{IEEEtran}
\bibliography{egbib}

\begin{IEEEbiography}[{\includegraphics[width=1in,height=1.25in,clip,keepaspectratio]{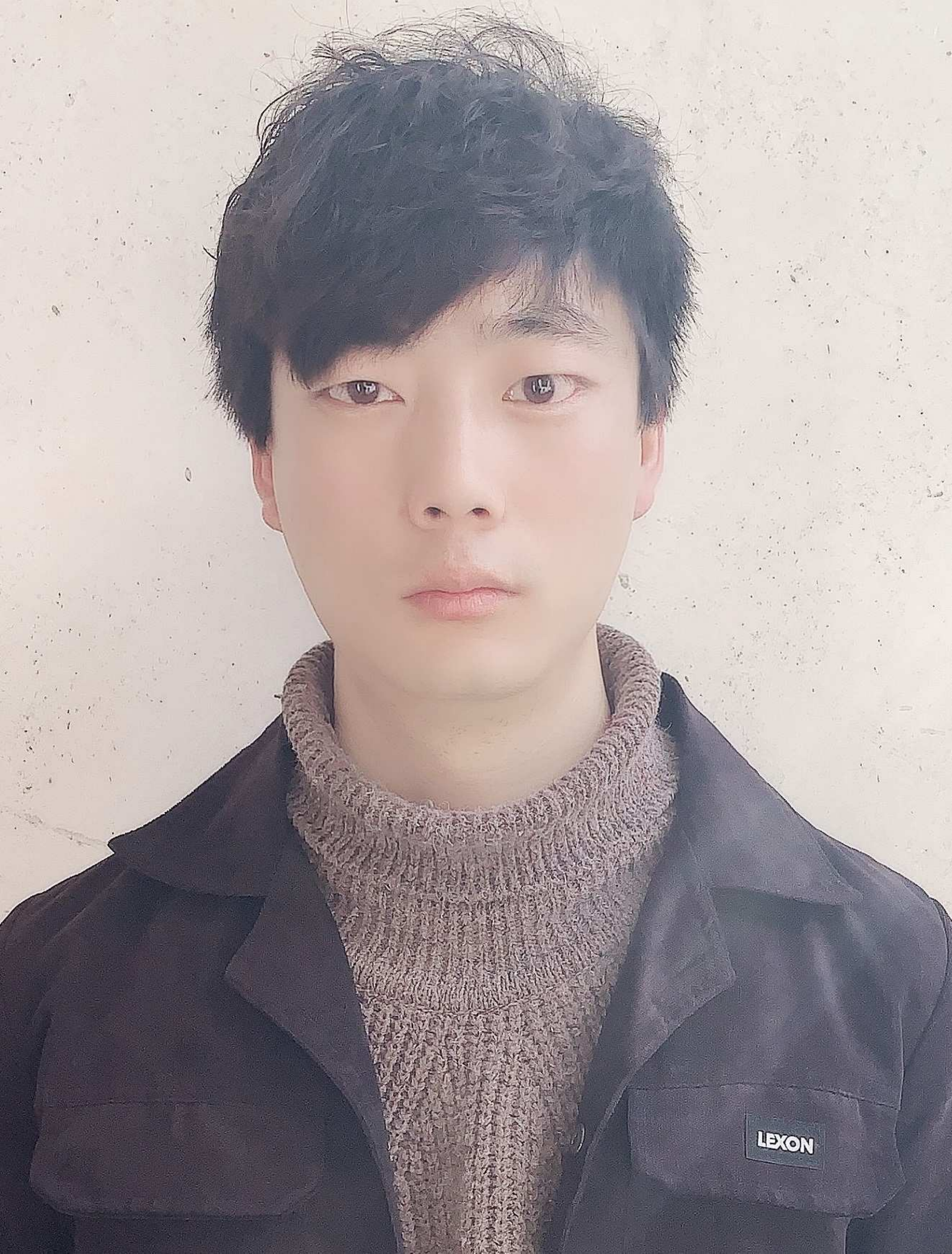}}]
{Xiang Zhang}
received the MS degree in computer science from Northwest University, Xi'an, China, in 2018. He is now a PhD candidate at the same university. His research interests include statistical machine learning, large-scale visual recognition, plant species identification, and fine-grained semantic image segmentation.
\end{IEEEbiography}
\begin{IEEEbiography}[{\includegraphics[width=1in,height=1.25in,clip,keepaspectratio]{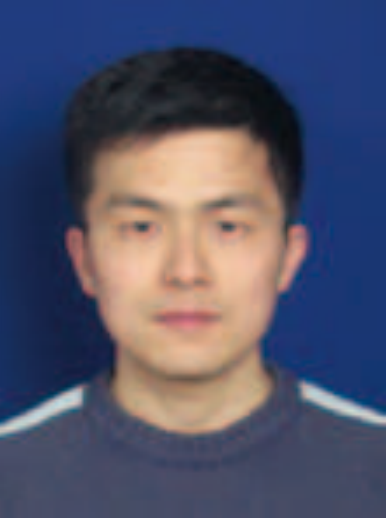}}]
{Wei Zhang}
received the PhD degree in computer science from Fudan University, China, in 2008. He is currently an associate professor with the School of Computer Science, Fudan University. He was a visiting scholar with the UNC-Charlotte, in 2016-2017. His current research interests include deep learning, computer vision, and video object segmentation.
\end{IEEEbiography}
\begin{IEEEbiography}[{\includegraphics[width=1in,height=1.25in,clip,keepaspectratio]{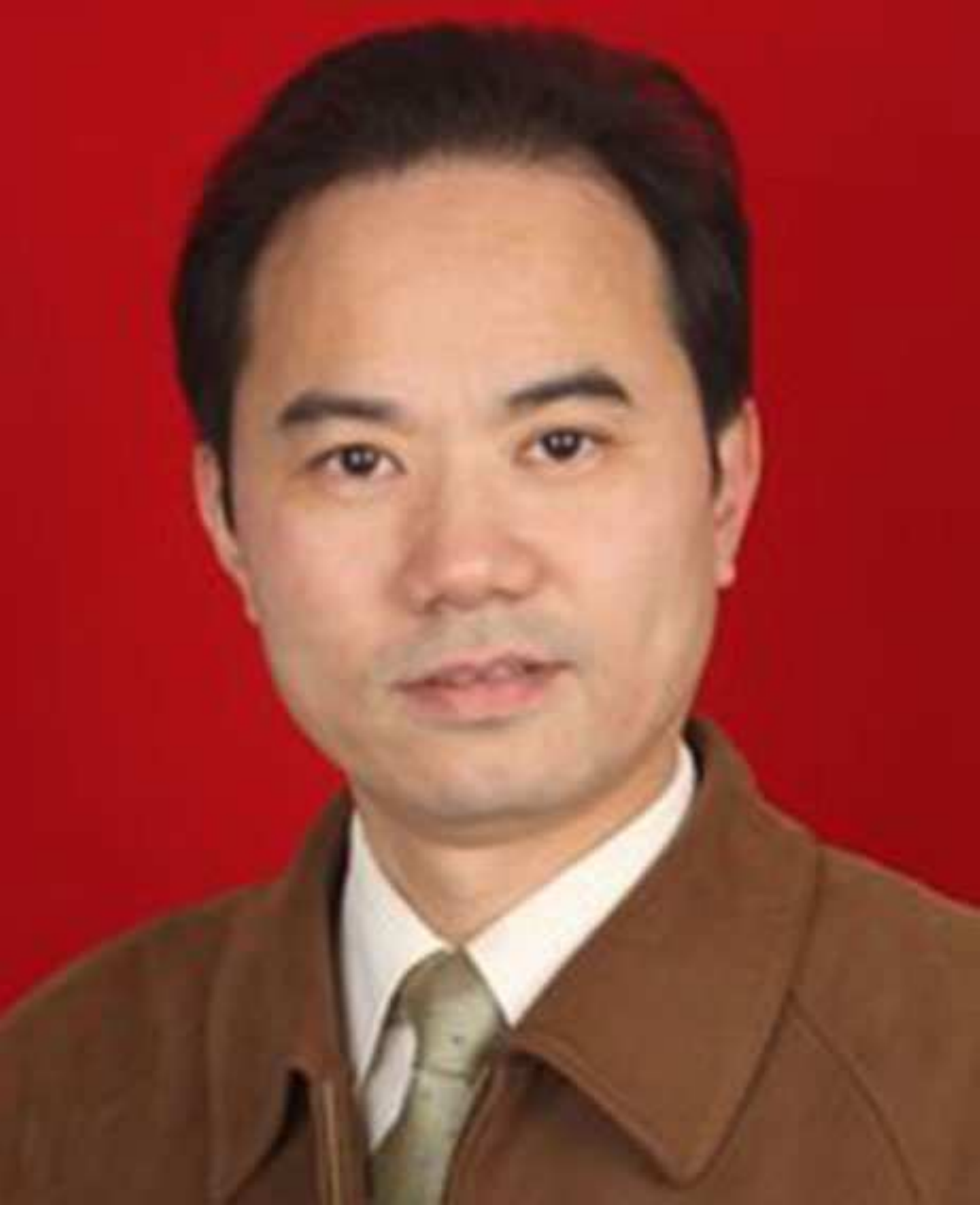}}]
{Jinye Peng}
received his MS degree in computer science from Northwest University, Xi'an, China, in 1996 and his PhD degree from Northwest Polytechnical University, Xi'an, China, in 2002. He joined Northwest Polytechnical University as Full Professor at 2006. His research interests include image retrieval, face recognition, and machine learning.
\end{IEEEbiography}
\begin{IEEEbiography}[{\includegraphics[width=1in,height=1.25in,clip,keepaspectratio]{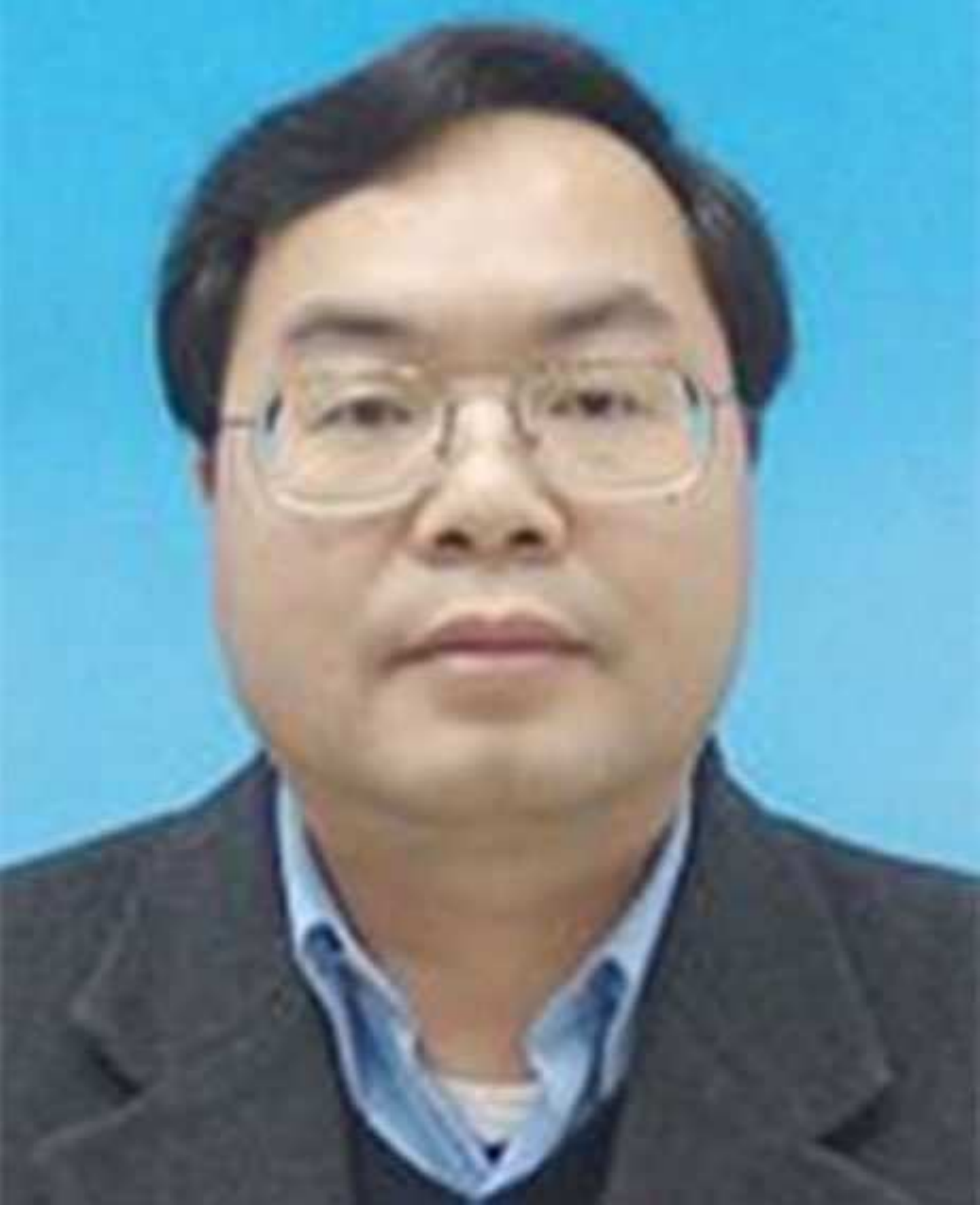}}]
{Jianping Fan}
is a professor at UNC-Charlotte. He received his MS degree in theory physics from Northwest University, Xi'an, China in 1994 and his PhD degree in optical storage and computer science from Shanghai Institute of Optics and Fine Mechanics, Chinese Academy of Sciences, Shanghai, China, in 1997. He was a Researcher at Fudan University, Shanghai, China, during 1997-1998. From 1998 to 1999, he was a Researcher with Japan Society of Promotion of Science (JSPS), Osaka University,
Japan. From 1999 to 2001, he was a Postdoc Researcher in the Department of Computer Science, Purdue University, West Lafayette, IN. His research interests include image/video privacy protection, automatic image/video understanding, and large-scale deep learning.
\end{IEEEbiography}




\end{document}